\documentclass[pmlr,twocolumn]{jmlr}
   
\usepackage{longtable}

\usepackage{booktabs}
\usepackage[load-configurations=version-1]{siunitx} 

\usepackage{comment}
\usepackage{hhline}
\usepackage{makecell} 
\usepackage{multirow}

\theorembodyfont{\upshape}
\theoremheaderfont{\scshape}
\theorempostheader{:}
\theoremsep{\newline}

\title[Instability in clinical risk stratification models]{Instability in clinical risk stratification models\\ using deep learning}

  \author{
  \Name{Daniel Lopez-Martinez\nametag{\thanks{Corresponding author.}}} \Email{dlmocdm@google.com}\\
  \Name{Alex Yakubovich} \Email{alexyak@google.com}\\
  \Name{Martin Seneviratne} \Email{martsen@google.com}\\
  \Name{Adam D. Lelkes} \Email{lelkes@google.com}\\
  \Name{Akshit Tyagi} \Email{akshittyagi@google.com}\\
  \Name{Jonas Kemp} \Email{jonasbkemp@google.com}\\
  \addr Google Research, Mountain View, CA, USA
  \AND
  \Name{Ethan Steinberg} \Email{ethanid@stanford.edu}\\
  \Name{N. Lance Downing} \Email{ldowning@stanford.edu}\\
  \Name{Ron C. Li} \Email{ronl@stanford.edu}\\
  \Name{Keith E. Morse} \Email{keith.morse@stanford.edu}\\
  \Name{Nigam H. Shah} \Email{nigam@stanford.edu} \\
  \addr Department of Medicine, Stanford University, Stanford, CA, USA
  \AND
  \Name{Ming-Jun Chen} \Email{mingjunchen@google.com} \\
  \addr Google Research, Mountain View, CA, USA
  }

\begin{document}

\maketitle

\begin{abstract}
While it has been well known in the ML community that deep learning models suffer from instability, the consequences for healthcare deployments are under characterised.
We study the stability of different model architectures trained on electronic health records, using a set of outpatient prediction tasks as a case study. 

We show that repeated training runs of the same deep learning model on the same training data can result in significantly different outcomes at a patient level even though global performance metrics remain stable. 

We propose two stability metrics for measuring the effect of randomness of model training, as well as mitigation strategies for improving model stability. 

\end{abstract}

\section{Introduction} \label{sec:intro}

Machine learning methods have shown great promise in a wide range of healthcare applications, from diagnosing conditions from imaging, to personalizing care, to improving administrative workflows (see e.g.~\cite{Davenport2019-qu}).
One particular use case that has attracted a lot of research interest is \emph{risk stratification}; i.e., assigning each patient a risk status using a machine learning model trained to predict a patient's individual risk of some adverse event.
Previously studied risk prediction tasks cover a range of inpatient and outpatient risks, from acute inpatient predictions such as acute kidney injury~\citep{Mohamadlou2018-hp} to longer term outpatient predictions such as readmissions~\citep{Huang2021-ol}.

These risk stratification use cases are typically motivated by the promise of improving overall patient care by helping clinicians prioritize the highest risk patients. Therefore, they can be used to inform the allocation of scarce and valuable resources, such as a care manager's time, among a group of patients. Whenever algorithmic tools are used for resource allocation, especially in such an important and sensitive area as healthcare, ensuring the robustness and fairness of the algorithmic decision-making process is of utmost importance.

In this paper, we want to draw the community's attention to a particular source of concern in this respect: model instability. As in many other areas of machine learning applications, deep learning methods often demonstrate state-of-the-art performance for risk prediction tasks. It is well known in the machine learning community that deep learning models are more sensitive to random phenomena during model training, and their outputs are often less stable than those of traditional, simpler models. However, the practical implications of model instability for risk stratification haven't been well studied.

As a case study for this phenomenon, we train feed-forward neural network model on EHR data to predict all-cause emergency admissions, along with a set of disease-specific admission events.
These models are trained an data from the EHR of an academic medical center.
Importantly, we evaluate the impact of instability in a pragmatic fashion, by simulating the prospective deployment setting of the model---namely, as a method to select high-risk patients for a care manager to contact. 
We compare deep learning  (DL) models against simpler logistic regression (LR) models on both performance and stability metrics. We conclude by proposing an ensembling approach to mitigate against DL instability and suggest directions for future methodological work.

While we used unplanned hospital admission  predictions as our case study, the concept of model stability and our methods to evaluate it are generalizable to any other models where the ordinal nature of the risk scores is important---essentially any clinical workflow where resources (e.g. capacity to intervene or follow up) are restricted.

\section{Related work} \label{sec:relatedwork}

Risk stratification models have been built for a number of prediction tasks, such as forecasting mortality \citep{Luo2022-nw}, patient deterioration \citep{Mann2021-hr}, unplanned hospital admissions \citep{King2022-aw}, or the need for medical procedures \citep{Lopez-Martinez2022-in}. However, the  performance of these models is usually evaluated using traditional metrics of predictive performance, such as the  area  under  the  curves  (AUC)  of the  receiver  operating  characteristic  (ROC-AUC)  and  the  precision-recall  curve  (PR-AUC). Although having merit, recent work has highlighted the need for additional metrics to evaluate models.

For example, \cite{Misic2021-bs} has noted that traditional metrics lack interpretability and  provide an incomplete picture of the overall benefit of the model as they do not take into account how providers may use them. For example, if used to allocate scarce medical resources, resource constraints should inform model evaluation. \cite{King2022-aw} has also noted that building a model for real-world implementation involves several additional challenges to those encountered when simply optimising a model's technical performance, such as evaluating the drift in model performance over time in deployment simulations. Model fairness~\citep{Rajkomar2018-ol, Mhasawade2021-aq} should  also be key aspect of model evaluation, to avoid disparities in resource allocation. Therefore, if risk-stratification models are to be used operationally, model evaluation should go beyond the traditional metrics.

Reproducibility is also a key aspect of the assessment of a model, and a minimal prerequisite for any machine learning model applied to human health (\cite{McKinney2020-xe, Beam2020-jh, Liu2022-nn}). Lack of reproducibility undermines the credibility of a prediction, and diminishes user trust in the machine learning model. Moreover, regulatory agencies are implementing rules governing medical use of machine learning (\cite{Vollmer2020-fr}), and may reasonably expect being able to duplicate the results of model's evaluation using the same elements as were used in the prior study.

Unfortunately, even models trained with the same training algorithm and dataset can have different model parameters and yield different predictions. This is due to nondeterminism during  training  \citep{Nagarajan_undated-qc, Summers2021-cn, Madhyastha2019-yi}, which includes:
\begin{itemize}
    \item \textbf{Parameter initialization} Prior to training, the weights of a NN are initialized randomly (often from a Gaussian or uniform distribution centered at zero) ~\citep{Glorot2010-ji}.
    \item \textbf{Stochastic optimization} NNs are trained with stochastic gradient descent (SGD) or other similar stochastic optimization techniques, which typically sample small batches of examples from the training set at random in each iteration.
    \item \textbf{Stochastic regularization layers} Regularization techniques such as dropout or noisy activation functions ~\citep{Gulcehre2016-wn}.
\end{itemize}

In addition to randomness due to stochastic training algorithms, the hardware and software tooling used for model training can also result in differences across models trained with the same data and algorithm. These often trade determinisim and precision for speed, introducing nondeterminism e.g. by the random completion order of parallel tasks~\citep{Zhuang2022-nj}.

In the EHR literature, model stability issues have already been identified. \cite{DAmour2020-ih} and \cite{Dusenberry2020-ph} have noted that changing random seeds in training led to substantially different test set predictions. Therefore, \cite{Dusenberry2020-ph} highlighted that model uncertainty is a key factor one should pay attention to while adopting a DL approach to predict events from EHR data. Despite these insights, the instability of deep learning models with EHR data, and specifically its impact in the real-world use of risk stratification models,  remains poorly  understood.
For example, DL training instability has been shown to lead to variability in model fairness \citep{Qian2021-af}, but this has not been studied in clinical models.

\section{Methods} \label{sec:methods}

\subsection{Dataset} \label{sec:dataset}


We used a limited dataset provided by the STAnford medicine Research data Repository (STARR-OMOP) \citep{Datta2020-rw},  which consists of Stanford’s EHR data in the Observational Medical Outcomes Partnership (OMOP) Common Data Model (CDM). Data from Lucile Packard Children’s Hospital (LPCH), which is a pediatric institution, was excluded from this study. 
The dataset was utilized in accordance with HIPAA and Stanford Medicine Research IT processes, including IRB review.

To leverage existing FHIR-based modeling infrastructure, we manually mapped the dataset from the OMOP CDM to the HL7 Fast Healthcare Interoperability Resources (FHIR) standard (version R4) (\cite{Bender2013-oh}).

\subsection{Model development}

To explore the stability of risk stratification methods, we implemented 4 outpatient deterioration tasks as a case study. 

\subsubsection{Task definition}
We predicted the 30-day risk of all-cause emergency hospitalizations and three condition-specific potentially preventable unplanned hospitalizations: chronic obstructive pulmonary disease (COPD), diabetes and heart failure (HF).

Emergency admissions were identified by examining the class and admit source of the medical encounter. The 3 disease-specific unplanned hospitalizations were defined using the inclusion and exclusion criteria of the Agency for Healthcare Research and Quality’s Patient Quality Indicators (PQIs) for ambulatory care-sensitive conditions \citep{Agency_for_Healthcare_Research_and_Quality2016-ee, Agency_for_Healthcare_Research_and_Quality2018-lb}, and identified by examining the diagnosis and procedure codes corresponding to the PQI definition \citep{Quan2005-wv}.

\subsubsection{Label extraction}

Following \cite{Rajkomar2018-kv}, we represented each patient's history as an ordered timeline of FHIR resources. As in \cite{Lopez-Martinez2022-in}, for each patient we generated a prediction on the first day of every month within the study period, if and only if the patient had at least one FHIR resource within the preceding year. Each example in our dataset corresponded to one such monthly prediction: we included all resources occurring prior to the prediction time as features, and we computed a binary label denoting whether the patient had an emergency admission within the first 30 days following the prediction time.

\subsubsection{Feature extraction}

For each patient, we extracted information about the demographics, conditions, procedures, medications, observations and encounters.

The demographics features were gender and age. 
The condition-based features contained all International Classification of Diseases (ICD) codes associated with each encounter. The procedure-based features included all Current Procedural Terminology (CPT) codes, as well as ICD procedure codes. The medication-based features contained RxNorm codes \citep{Liu2005-tw} for medication requests. The observation-based features included indicators for measurements of patients' vital signs and lab results and were constructed by concatenating the observation code (LOINC \citep{McDonald2003-nf} and SNOMED \citep{Cornet2008-nh} codes) and the unit of measurement. The encounter-based feature was the class of encounter.

Following \cite{Lopez-Martinez2022-in},  we bucketed the feature values into one of four disjoint time intervals indicating how long before the prediction time that feature value occurred: less than 30 days before, 30-90 days before, 90 days-1 year before, or 1-10 years before. The demographic features (age and gender) were not bucketed. Each feature was represented as a multi-hot vector.

\subsubsection{Model architecture}

Using the aforementioned features, we designed independent logistic regression (LR) and feedforward neural network (FNN) machine learning models for each of the 4 tasks. The FNN consisted of three hidden layers of 128, 64, and 32 units respectively, with ReLU activations and dropout after each hidden layer, followed by a final prediction layer. Weights for the FNN model were randomly initialized using Glorot uniform initialization ~\citep{Glorot2010-ji} on each training run; weights for the LR model were simply initialized to zero. All models were trained using Adam ~\citep{Kingma2014-ur}, a stochastic gradient-based optimizer method, with batch size 512. Predictions were rendered on the 1st day of every month on those patients who had at least one health interaction within the preceding year and were aged 18 years or older.

All models were implemented and trained with TensorFlow in Python \cite{Abadi2016-it}. Model hyperparameters, including L1/L2 regularization coefficients, dropout rates, and the parameters of an exponential learning rate decay, were tuned using Google Vizier \cite{Golovin2017-do} to minimize loss on the validation set.

\subsubsection{Model calibration} \label{sec:modelcalibration}

While it is desirable for our models to output well-calibrated risk probabilities, our LR and FNN models, like  most supervised learning methods, produce scores that can be used to rank the predictions from highest to lowest risk, but are not accurate probability estimates.

To address this limitation,  we used isotonic regression~\citep{Zadrozny2002-zh, Niculescu-Mizil2005-yf} to calibrate these models and transform ranking scores into probability estimates. Specifically, we fit a non-parametric isotonic (monotonically increasing) step-wise function $m$ to the original model scores $f_i$ such that $y_i = m(f_i) + \epsilon_i$, where $y_i$ denotes the true targets. Isotonic regression works by finding the isotonic function $\hat{m}$ such that:
\begin{equation}
    \hat{m} = \arg \min _z \sum (y_i - z(f_i))^2
\end{equation}
subject to $z(f_i) \ge z(f_j)$ whenever $f_i \ge f_j$.

All risk predictions reported in this papers are calibrated probabilities obtained from after applying isotonic regression.

\subsection{Model evaluation}

Model evaluation was done on an fixed independent held out test set comprising a 10\% random selection of the available patients. No patients in the test set were used for model training or validation. 

Performance was evaluated using two approaches. In the first one, we aggregated all test set predictions to compute the ROC-AUC and  PR-AUC.  Note that given the low prevalence of triggers with positive labels, the PR-AUC metric is more informative than ROC-AUC~\citep{Saito2015-oi}. In the second approach, we simulated the real-word deployment of the model by evaluating the test set predictions of each month independently (from January 2009 to December 2021), and then averaging these monthly metrics. This last evaluation approach allows us to introduce capacity constraints by examining the top K highest risk patients each month only.

\subsection{Stability analysis} \label{sec:methods_stability}
To quantify  model stability, in addition to the standard deviation of the ROC-AUC and PR-AUC metrics, we implemented the following metrics, which are computed using $N=10$ model runs with exactly the same training data, configurations and model hyperparameters, but different random seeds. 

\subsubsection{Top K Jaccard index}

In risk stratification methods, users may care about the $K$ highest-risk patients, as they are often limited in the amount of patients they can act upon. Therefore, it is reasonable to expect that the list of top $K$ patients should be the same across different models. 

To assess the similarity of the $N$ patient lists, we use the pairwise Jaccard index which  calculates the average Jaccard index across all pairwise comparisons. The Jaccard index is given by
\begin{equation}
    j({X^K_i, X^K_j}) = \frac{|X^{K}_i \cap X^{K}_j |}{|X^{K}_i \cup X^{K}_j |} 
\end{equation}
where $X^K_i$ and $X^K_j$ denote the top $K$ predictions in models $i$ and $j$ respectively.

For $N$ models, there exist ${N}\choose{2}$ pairwise comparisons. Therefore, the top $K$ pairwise Jaccard index is defined as follows:
\begin{equation}
    J = \frac{1}{ {{N} \choose{2}} } \sum_{i<j} j({X^K_i, X^K_j})
\end{equation}

The choice of $K$ depends on the specific use case.

\subsubsection{Rank correlation coefficient}

Risk scores are often used to prioritize patients, so that those with higher risk scores are acted upon earlier. In these scenarios, the ordinal nature of the patient lists is important. Users would expect models to produce similarly ranked patient lists.

Kendall's $\tau$ is commonly used to compare the ordinal correlation between two ranked lists~\citep{Kendall1938-yo, Kendall1945-lt}. 

It is based on computing the number of concordant ($|C|$) and discordant ($|D|$) pairs:
\begin{equation}
\tau = \frac{|C|-|D|}{{ {N}\choose{2} }}
\end{equation}
where a pair $((x_i, y_i),(x_j, y_j))$ is defined to be concordant if the sort order of the $x$'s is the same as the sort order of the $y$'s, otherwise it is discordant. 

We also consider a weighted version of Kendall's tau with hyperbolic weighting, in which correlations between top ranked patients have more influence on the metric (\cite{Vigna2015-nc}).

In this formulation, a datapoint of rank $r$ is assigned a weight of ${(r+1)}^{-1}$. The weight of exchanging two datapoints with ranks $r$ and $s$ is the sum of their weights, i.e. $ {(r+1)}^{-1} + {(s+1)}^{-1}$.

Weighting captures the assumption that the relative ordering of very low-risk patients is less important since interventions are typically aimed at high-risk patients.

\subsection{Ensembling} \label{sec:methodsensemble}

Ensembling is a popular approach for improving prediction performance across many different tasks, and it also reduces run-to-run variablity.
Averaging the outputs of multiple models reduces the random noise which might be causing the unnecessary variance in the predictive distribution~\citep{Huang2017-ui,Xie2013-oa}. It has also been shown to give better-calibrated estimates for uncertainty in the predictive distributions~\citep{Lakshminarayanan2016-oj}.
In our experiments we use a simple arithmetic mean of model outputs as a baseline; more complex averaging schemes could potentially yield further improvement.

\section{Results} \label{sec:methods}

\subsection{Dataset} \label{sec:methodsdataset}

\begin{table*}[h] \centering \small
\begin{tabular}{|l|r|r|rr|rr|}
\hline
\multicolumn{1}{|c|}{\multirow{2}{*}{Task}}  &  \multicolumn{1}{c}{\multirow{2}{*}{\# predictions}}   & \multicolumn{1}{|c|}{\multirow{2}{*}{\#  positives}}   & \multicolumn{2}{c|}{ROC-AUC}  & \multicolumn{2}{c|}{PR-AUC}   \\ \cline{4-7} 
\multicolumn{1}{|c|}{}                       & \multicolumn{1}{c}{}                                       & \multicolumn{1}{|c|}{}                                    & \multicolumn{1}{c|}{LR}       & \multicolumn{1}{c|}{FNN} & \multicolumn{1}{c|}{LR} & \multicolumn{1}{c|}{FNN} \\ \hline \hline 
Emergency       & 60,259,528        & 104,123 (0.17\%)      & \multicolumn{1}{r|}{0.9052} & \textbf{0.9116} & \multicolumn{1}{r|}{0.0734} & \textbf{0.0858} \\ \hline
COPD            &  5,759,224        &   1,228 (0.02\%)      & \multicolumn{1}{r|}{0.8894} & \textbf{0.9234} & \multicolumn{1}{r|}{0.0047} & \textbf{0.0069} \\ \hline
Diabetes        &  5,034,065        &     701 (0.01\%)      & \multicolumn{1}{r|}{0.8273} & \textbf{0.8919} & \multicolumn{1}{r|}{\textbf{0.0177}} & 0.0057 \\ \hline
Heart failure   &  2,161,112        &   2,369 (0.11\%)      & \multicolumn{1}{r|}{0.8717} & \textbf{0.8789} & \multicolumn{1}{r|}{0.0160} & \textbf{0.0170} \\ \hline
\end{tabular}
\caption{Number of predictions and positive label prevalence in the development (train and validation) set, together with model performance on the aggregated test set predictions for single ($N=1$) LR and FNN models. For each metric, the best performing model is highlighted in bold. ROC: receiver operating characteristic; PR: precision-recall curve; AUC: area under the curve.}
\label{tab:aggregatedmetrics}
\vspace{-10pt}
\end{table*}

A total of 1,576,215 individuals were present in the dataset. Of these, 53.66\% were females. White was the most predominant race (44\%), followed by asian (18.13\%), black (3.85\%), hispanic (2.02\%) and native american (1.12\%).

From these patients, we extracted a total of 66,977,919, 6,412,742, 5,588,138, and 2,400,145 predictions for the emergency, COPD, diabetes and heart failure tasks respectively. These predictions spanned a time period from January 1st 2009 to December 1st 2021. The differences in the number of predictions for each task can be explained by the different cohort inclusion criteria for prediction rendering. 

Using this data, we generated training, validation and test sets by randomly selecting 80\%, 10\% and 10\% of patients respectively. Positive label prevalence in the development set (training and validation sets) for the 4 tasks is shown in Table \ref{tab:aggregatedmetrics}.

\subsection{Model performance}

Table \ref{tab:aggregatedmetrics} shows the performance of the LR and FNN models for all 4 tasks in terms of ROC-AUC and PR-AUC. Note that given the imbalance of the data (positive labels were overwhelmingly underrepresented),  PR-AUC is more informative (\cite{Saito2015-oi}).

The best performance was obtained by the FNN model architecture, with the exception of the diabetes task, where it was outperformed by the linear model.  In both cases, the emergency task achieved the best performance, and did significantly better than the 3 remaining tasks. This may be explained by the significantly higher label prevalence of the emergency task.

\subsection{Stability analysis} \label{sec:resultsstability}

In order to conduct the model stability analysis described in Sec.\ref{sec:methods_stability}, for each task and model architecture we trained $N=10$ models using the same model hyperparameters and training data. Differences in model parameters and predictions across model runs may only be explained by nondeterminism in the training algorithm and hardware. Note that the choice of $N$ was based on visual inspection of the convergence of the stability metrics as we increased $N$. We determined that increasing it beyond $N=10$ would not have provided any additional value.

\begin{figure}[ht!] \centering
\includegraphics[width=0.95\linewidth]{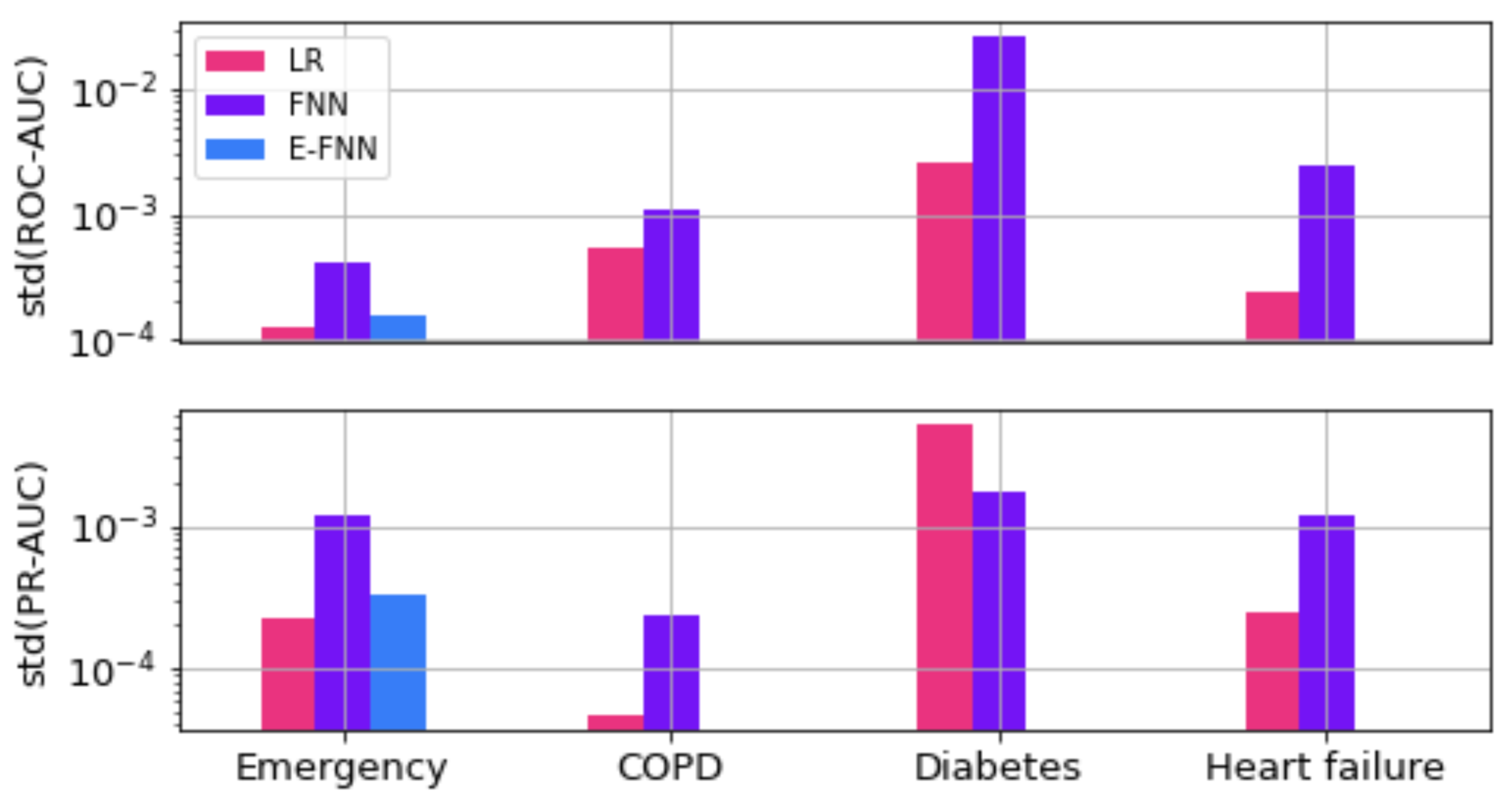}
\caption{Standard deviation of the ROC-AUC and PR-AUC metrics.}
\label{fig:std_traditional_metrics}

\includegraphics[width=0.94\linewidth]{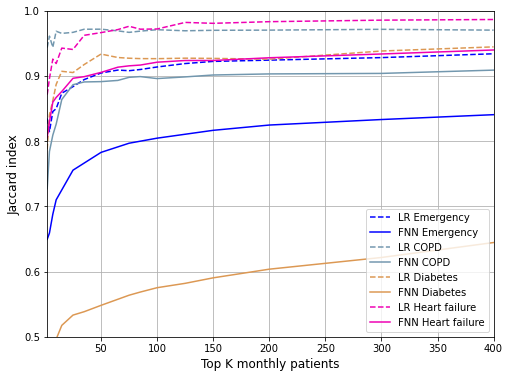}
\caption{Average pairwise Jaccard index $J(K)$ for all models. }
\label{fig:jaccard}

\includegraphics[width=0.96\linewidth]{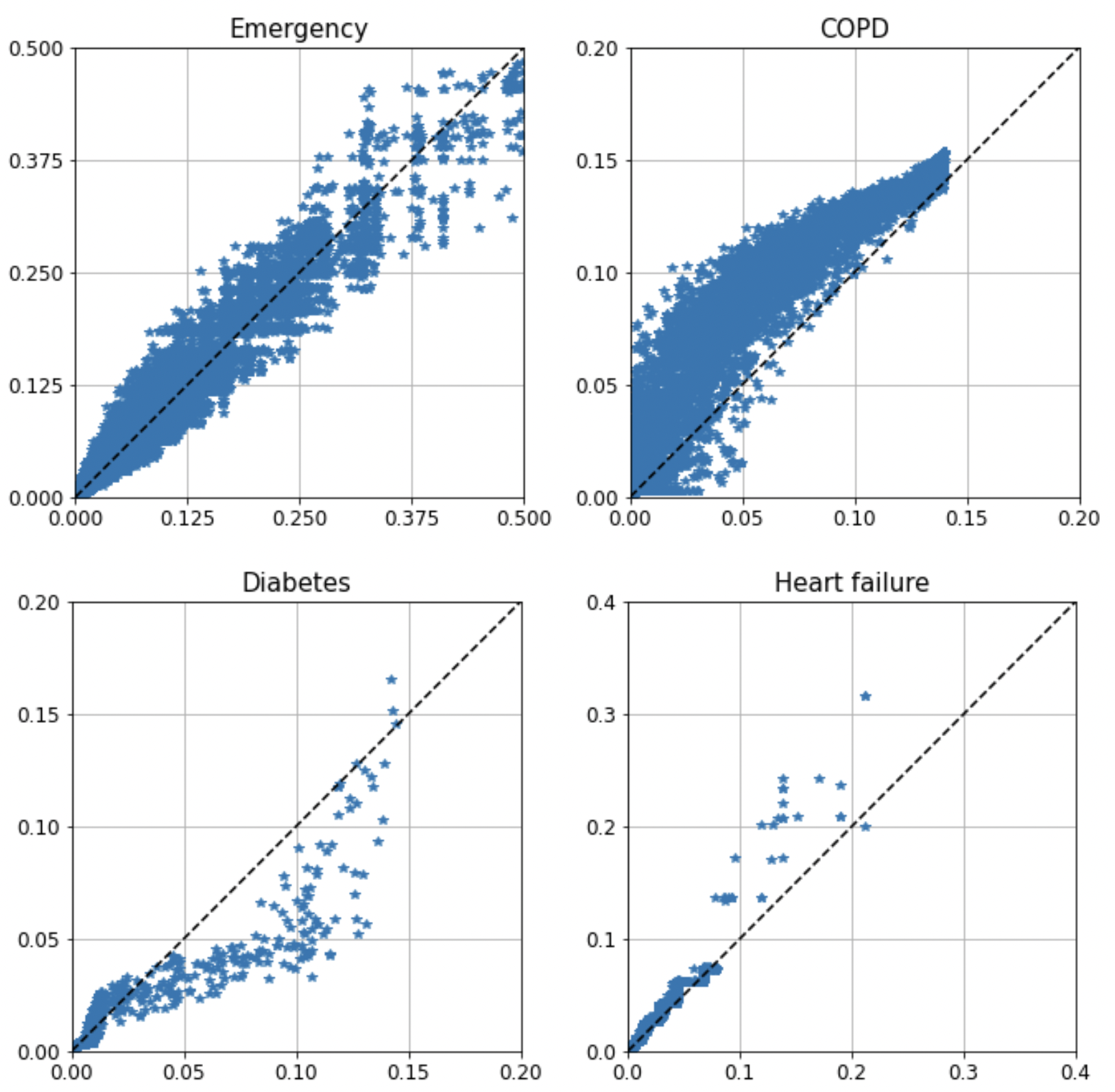}
\caption{Comparison of calibrated risks produced by two FNN models.}
\label{fig:pairwise_comparison}
\vspace{-20pt}
\end{figure}

First, we computed the standard deviation of the ROC-AUC and PR-AUC metrics across the $N$ model runs. This is depicted in Fig.\ref{fig:std_traditional_metrics}, which shows higher variability for the FNN architecture, with the exception of the PR-AUC metric of the diabetes model. Also, note that all  PR-AUC standard deviation metrics are smaller than $10^{-2}$, and these are difficult to contextualize and interpret. Hence, the standard deviations of the ROC-AUC and PR-AUC metrics may be inadequate for the assessment of model stability.

A better assessment of model stability may be provided by the  pairwise Jaccard index $J(K)$, which is shown in Fig.\ref{fig:jaccard} for different values of $K$. By illustrating average differences in top $K$ patient lists, it provides a more interpretable stability assessment. Of note is that the linear architecture achieves the best stability when compared to the FNN one, for all tasks and values of $K$. In addition to this, note that $J(K)$ decreases inversely with $K$. This suggests that stability is less of an issue for large values of $K$, but when resources are constrained and $K$ represents a small subset of the patient population, patient lists may be impacted by instability.

Note that given that the test set is comprised of a random sample of 10\% of the patients available in the STARR-OMOP dataset, a top K value of e.g. $K=100$ in the test set would correspond to 1,000 monthly patients in the entire dataset. 

A different insight into model stability is provided by Kendall's $\tau$ rank correlation coefficient, which, unlike the Jaccard index, takes into account the relative ordering of the patient risk scores. We consider both the standard Kendall's $\tau$ and one with hyberbolic weighting, where low-risk patients are downweighted (\cite{Vigna2015-nc}). We observed that the FNN model resulted in lower values of both weighted and unweighted $\tau$ with the exception of the unweighted $\tau$ of the COPD task, which confirms the previous insights that the LR model is more stable. In addition to this, we noted that using hyperbolic weighting resulted in larger values of $\tau$, suggesting that ordinal relationships are more preserved for the highest ranked patients.

\begin{table}[] \footnotesize \setlength{\tabcolsep}{3pt}
\begin{tabular}{llll}
\hline
Task                           & Model  & Unweighted $\tau$ & Weighted $\tau$ \\ \hline
\multirow{3}{*}{Emergency}     & LR & 0.9647 (0.0027)             & 0.9912 (0.0029)            \\
                               & FNN    & 0.9314 (0.0026)              & 0.9817 (0.0032)    \\        
                               & EFNN    & 0.9796 (0.0017)             & 0.9947 (0.0008)            \\ \hline
\multirow{2}{*}{COPD}          & LR & 0.9267 (0.0214)              & 0.9864 (0.0049)            \\
                               & FNN    & 0.9540 (0.0013)              & 0.9831 (0.0020)            \\ \hline
\multirow{2}{*}{Diabetes}      & LR & 0.9060 (0.0188)              & 0.9797 (0.0037)            \\
                               & FNN    & 0.5883 (0.0700)              & 0.8312 (0.0704)            \\ \hline
\multirow{2}{*}{HF} & LR & 0.9870 (0.0010)              & 0.9950 (0.0007)            \\
                               & FNN    & 0.8847 (0.0197)              & 0.9704 (0.0038)            \\ \hline
\end{tabular}
\caption{Average pairwise Kendall's rank correlation coefficient ($\tau$), and standard deviations, with uniform  and hyperbolic weighting.} 
\label{tab:kendalltau}
\vspace{-16pt}
\end{table}

Finally, we explore the differences in calibrated risk scores. 
Fig.\ref{fig:pairwise_comparison} compares two randomly selected models for each task. Of note is that an individual's calibrated risk score can vary significantly due to training variability. For the emergency model, across all pairwise comparisons we observed that the highest ranked patients experienced the largest variability (see Fig.\ref{fig:pairwise_comparison}, top left, for an example). This behavior was occasionally present in the disease-specific tasks, but the risks of the mid-ranked patients varied more more as illustrated in the comparison of the COPD and diabetes tasks.

\subsection{Ensemble}

We trained $N=10$ averaging ensembles, each one comprised of $M=10$ models whose final outputs are averaged. Due to the  cost of training  $N\times M$ models, we focused our analysis on the FNN emergency model only. 

\begin{figure}[h!] \centering
\includegraphics[width=0.95\linewidth]{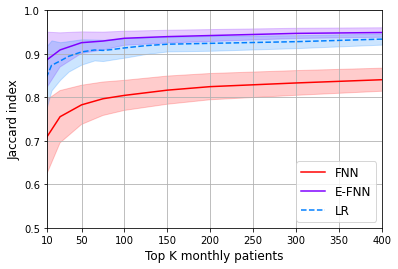}
\caption{Average pairwise Jaccard index $J(K)$ for the LR, FNN and E-FNN emergency models. The shadow represents one standard deviation computed over all prediction times.}
\label{fig:emergency_ensemle}

\includegraphics[width=1.00\linewidth]{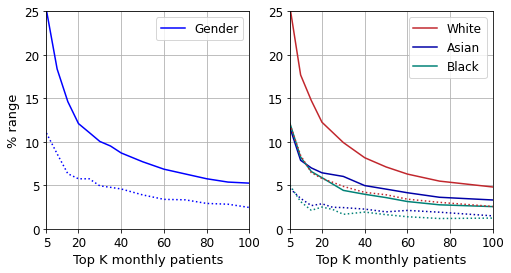}
\caption{Range in gender (left) and race (right) composition of the top K patients for the $N$ runs of the FNN (solid lines) and E-FNN (dotted lines) emergency models.}
\label{fig:fairness}
\vspace{-17pt}
\end{figure}

Ensembling did not significantly impact the average ROC-AUC, which was 0.9111 for both the FNN and ensembled FNN (E-FNN), but did improve the average PR-AUC from 0.0843 to 0.0848, and the variance of these AUC metrics, as shown in Fig.\ref{fig:std_traditional_metrics}. Kendall's $\tau$ (Table \ref{tab:kendalltau}) and the Jaccard index ( Fig.\ref{fig:emergency_ensemle}) also improved significantly,  surpassing the linear architecture in stability.

\subsection{Fairness analysis}
Finally, we investigate whether the stability issues discussed in Sec.\ref{sec:resultsstability} impact the fairness of the model, specifically, whether it differentially impacted the gender and race representation in the monthly lists of top K highest-risk patients. 

For each patient and prediction time, we computed the range in representation across $N=10$ models. This is illustrated in Fig.\ref{fig:fairness} for the FNN emergency model and different values of $K$. It shows that in addition to leading to different patient selections and individual risk scores, instability can also lead to gender and race differences in patient selection. Note that while smaller differences were observed for the Asian and Black race categories, that is explained by their reduced prevalence in the dataset (see Sec.\ref{sec:methodsdataset}). 

Finally, we also noted that the E-FNN reduced both gender and race variability. For example, for $K=30$, the range in gender across the $N$ FNN model runs was 10.04\%, whereas the range across the $N$ E-FNN model runs (where each E-FNN model was derived from $M=10$ independent FNN models) was 4.94\%. Similarly, the range in  race composition decreased with the ensemble from 9.92\% to 4.88\%, 6.02\% to 2.44\%, and 4.42\% to 1.67\% for white, asian and black races respectively.

\section{Discussion} \label{sec:discussion}

This work examined the instability of risk stratification models resulting from nondeterminism in the training process. Using a set of  unplanned hospital admission prediction tasks as a case study, we confirmed previous observations that repeated training runs can lead to significantly different predictions at a patient level even if traditional model accuracy metrics remain unchanged.

We have proposed two similarity metrics to quantify this run-to-run variability among risk stratification models in a way that accounts for real-world usage under resource constraints. The first one, the Jaccard index, evaluates the consistency in patients selected for intervention when the model is used to allocate limited resources. The second one, Kendall's rank correlation coefficient, evaluates whether the ordinal relationships are preserved; therefore, it is broadly useful to evaluate models that prioritize patients for intervention. These stability metrics can also be used to select the most stable models.

We have also shown that the observed instability can lead to variance in the representation of subgroups among the top K predictions, which may have implications for equitable allocation of scare resource and could in turn diminish trust in the models.

Finally, we have evaluated a simple strategy (model ensembling) to mitigate model instability. There are other potential strategies, such as implementing techniques for addressing data imbalance (e.g. synthetic minority oversampling or weighted loss functions), decaying the learning rate, replacing the ReLU activation function with SmeLU~\citep{Gil2020} etc., that can also improve the model stability. However, producing a throughout review of strategies to improve stability is not in the scope of this work; instead, we want to provide an example to show how stability issue can be mitigated. 
Our averaging ensemble improved all stability metrics considered while preserving overall model performance.
However, this benefit of creating an ensemble comes with an extra cost of multiple model runs in memory and disk space. Future work should consider more efficient ensemble approaches, such as batch ensembling~\citep{Wen2020-ul}.

The acceptable level of instability will vary among different application use cases depending on a complex set of factors such as capacity to intervene, subgroup representation, and frequency of training runs; we do not make any claim as to whether any specific models we have examined are stable enough for particular applications.
However, we believe that model instability has been an overlooked factor that should receive more consideration when deciding whether to deploy a model to a healthcare setting. In order for potential users to be able to make informed decisions, researchers need to transparently report and discuss the stability of their proposed methods.  We hope that our work will lead to more transparency and that it will start a discussion about the importance of model stability in the research community.

\acks{We are grateful to Lan Huong Nguyen, Andreas Schagerer, Eric Loreaux and Bakhtiar Rehman
for their contributions to make this work possible.}

\bibliography{lopezmartinez22}

\end{document}